\documentclass{article} 
\usepackage{iclr2024_conference,times}
\usepackage{hyperref}
\usepackage{url}

\usepackage{amsmath,amsfonts,bm}
\usepackage{amssymb}
\usepackage{graphicx}
\usepackage{subcaption}
\usepackage{xcolor} 
\usepackage{soul} 
\usepackage{framed} 
\usepackage{color, colortbl} 
\usepackage{amsbsy} 
\usepackage{placeins} 
\usepackage{xr} 
\usepackage{makecell} 
\usepackage{hyperref} 
\usepackage{booktabs}

\newcommand{\tlt}{$TlT$}

\definecolor{keyword}{HTML}{0076BA}
\definecolor{grammodify}{HTML}{D9EAD3}

\newcommand\neweacl{}

\newcommand\neweswa{}
\newcommand\reveswa{}
\newcommand\reveswatwo{}

\title{Automatic Logical Forms improve fidelity in Table-to-Text generation}

\author{Iñigo Alonso, Eneko Agirre \\ 
HiTZ Basque Center for Language Technology - Ixa NLP Group \\ University of the Basque Country UPV/EHU \\
\texttt{\{inigoborja.alonso,e.agirre\}@ehu.eus}
}

\iclrfinalcopy 
\begin{document}

\maketitle

\begin{abstract}
Table-to-text systems generate natural language statements from structured data like tables. While end-to-end techniques suffer from low factual correctness (fidelity), a previous study reported fidelity gains when using \reveswa{manually produced graphs that represent the content and semantics of the target text called Logical Forms (LF)}. Given the use of manual \reveswa{LFs}, it was not clear whether automatic LFs would be as effective, and whether the improvement came from the implicit content selection \reveswa{in the LFs}. We present \tlt, a system which, given a table and \reveswa{a set of pre-selected table values}, first produces LFs and then the textual statement. We show for the first time that automatic LFs improve \reveswa{the quality of generated texts}, with \reveswa{a 67\% relative increase in fidelity} over a comparable system not using LFs. Our experiments allow to quantify the remaining challenges for high factual correctness, with automatic selection of content coming first, followed by better Logic-to-Text generation and, to a lesser extent, \reveswa{improved} Table-to-Logic parsing.
\end{abstract}

\section{Introduction}
\label{sec:intro}

Data-to-text generation is the task of taking non-linguistic structured input such as tables, knowledge bases, tuples, or graphs, and automatically producing factually correct\footnote{We use the terms factual correctness, faithfulness, and fidelity indistinctly.} textual descriptions of the contents of the input \citep{Reiter1997, Covington2001, Gatt2018}. \reveswa{Real-world applications include, among others, generating weather forecasts from meteorological data \citep{Goldberg1994}, producing descriptions from bioentgraphical information \citep{lebret-etal-2016-neural}, or generating sport summaries using game statistics \citep{Wiseman2017}. In these applications, the goal is to represent relevant information in the input data using natural language descriptions. Therefore, generating text that faithfully and accurately represents the underlying information in the source becomes critical.} \reveswa{It should be noted that the task is underspecified, in the sense that the same table may be described by multiple textual descriptions, all of them correct, as each one can focus on different, relevant subsets of the input data. This makes the use of manual evaluation of fidelity key to measure the quality of the generated text. Our work focuses on how to improve faithfulness automatically.} 

\reveswa{Various Data-to-Text approaches have emerged to address this challenge. Methods include leveraging the structural information of the input data \citep{Wiseman2017, Puduppully2019a, chen-etal-2020-kgpt}, using neural templates \citep{Wiseman2018}, or focusing on content ordering \citep{Puduppully2019b}.} Recent techniques \citep{Chen2020b, Chen2020a, Aghajanyan2021,Kasner2022} leverage large-scale pre-trained models \citep{Devlin2018}, 
and report significant performance gains \reveswa{in terms of fluency and generalization with respect to previous work that did not use such models}. 

However, these end-to-end systems struggle with fidelity \reveswa{as they are still susceptible to produce hallucinations, i.e. they generate text that, despite its fluency, does not describe in a faithful way the input data \citep{koehn-knowles-2017-six, maynez-etal-2020-faithfulness}.} 

\reveswa{In this context} \citet{Chen2020a} propose to reformulate Data-to-Text as a Logic-to-Text problem. \reveswa{Alongside the usual table information, the input to the language realization module in this approach also includes a tree-structured graph representation of the semantics of the target text called logical form (LF)}. \reveswa{Logical forms follow compositional semantics \citep{Carnap.1947} to formalize the underlying meanings represented in the target text. \reveswatwo{When provided alongside tables in this case, the meaning conveyed by LFs is related to a semantic context as defined in \citet{zhang1994integrated, 6714612}. In this case, the semantic context is given by the table.} An example of how LFs represent this meaning can be seen in Fig. \ref{fig:lf-example}. Although the LFs were applied to tables in this paper, the proposal could be easily extended to other Data-to-Text problems.}

\begin{figure}[t]
\includegraphics[width=\linewidth]{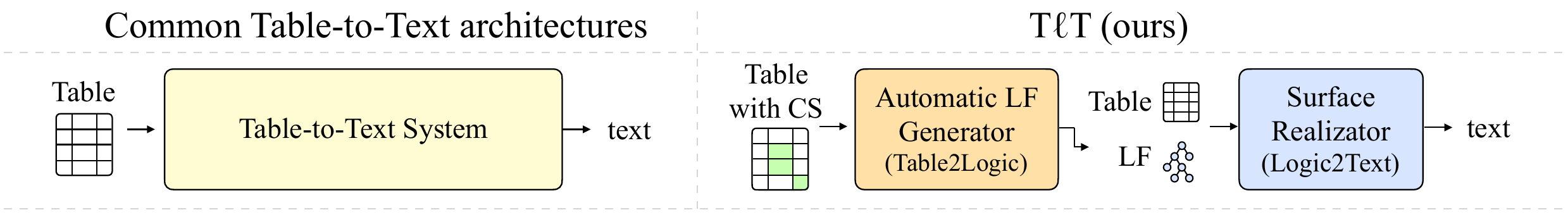}
\caption{Our proposed system to improve fideltiy, \tlt, (right) alongside a typical Table-to-Text architecture (left). 
}
\label{fig:tlt-comparison}
\end{figure}

\reveswa{With the use of manual LFs, \citet{Chen2020a} report an increase in factual correctness from 20\% to 82\% compared to a system not using LFs.} Manually produced LFs include, implicitly, a selection of the contents to be used in the description also referred as Content Selection (CS). \reveswa{Content Selection is the task of choosing the subset of the table that is to be communicated in the output \citep{duboue-mckeown-2003-statistical}. LFs inherently provide the content selection within themselves, and thus models based on manual LFs have an easier task and a lower probability of producing an unfaithful statement.} \reveswa{The main shortcoming of this approach is that the manual production of LFs is very costly and it is not realistic to expect table producers to add formal semantic representations such us LFs for each table that they produce. \cite{Chen2020a}} left two open research questions: Firstly, the improvement in faithfulness could come from the implicit content selection alone, casting doubts about the actual contribution of LFs. Secondly, it is not clear whether a system using automatic LFs would be as effective as a system based on manual LFs. \reveswa{Our goal is to answer these two questions.}

In this work we present \tlt~(short from Table-to-Logic-to-Text), a two-step model that produces descriptions by, first, automatically generating LFs (Table-to-Logic parsing), and then producing the text from those LFs (Logic-to-Text generation). Our model \neweacl{(see Figure \ref{fig:tlt-comparison}) allows Table-to-Text generation systems to leverage the advantages of using LFs without requiring manually written LFs. We separate the content selection process from the logical form generation step, allowing to answer positively to the open questions mentioned above with experiments on the Logic2Text dataset \citep{Chen2020a}. Although content selection alone improves results, the best results are obtained using automatic LFs, with noteworthy gains in fidelity compared to a system not using LFs. }
\neweacl{
Our results and analysis allow to estimate the impact in fidelity of the remaining challenges, with automatic content selection coming first, followed by better Logic-to-Text generation and to a lesser extent Table-to-Logic parsing. We also provide qualitative analysis of each step. }

All code, models and derived data are publicly available \footnote{\url{https://github.com/alonsoapp/tlt}}.

\section{Related Work}
\label{related-work}

Natural Language Generation from structured data is a long-established research line. Over time, multiple techniques have been developed to solve this task in different ways, such as leveraging the structural information of the input data \citep{Wiseman2017, Liu2018, Puduppully2019a, Rebuffel2020, chen-etal-2020-kgpt}, using neural templates \citep{Wiseman2018, Li2018} or focusing on content ordering \citep{Sha2018, Puduppully2019b, su-etal-2021-plan-generate}. \reveswa{The use of pre-trained language models \citep{Devlin2018, Radford2019} has allowed to improve text fluency compared to those early systems \citep{Chen2020b, Aghajanyan2021, Kasner2022}}; however, fidelity remains the main unsolved issue in all of the aforementioned systems.

\reveswa{A body of research has thus focused on improving factuality.} \citet{Matsumaru2020} remove factually incorrect instances from the training data. Other proposals take control of the decoder by making it attend to the source \citep{Tian2019}, using re-ranking techniques \citep{Harkous2020}, or applying constrains that incorporates heuristic estimates of future cost \citep{Lu2021}. Alternatively, \citep{Wang2020, Shen2020, Li2020} rely on heuristics, such as surface matching of source and target, to control generation. 

In a complementary approach \reveswa{to improve} factuality, \citet{Chen2020a} \neweacl{propose reformulating Table-to-Text as a Logic-to-Text problem. They incorporate a tree-structured representation of the semantics of the target text, logical forms (LF), along with the standard table information. The logical form highly conditions the language realization module to produce the statement it represents, significantly improving fidelity results. However, the logical forms in this work are manually produced by humans, which is unrealistic and greatly reduces the applicability of this solution in a real-world scenario. \reveswa{Our work builds on top of this approach, adopting LFs and proposing to generate them automatically based on table data alone, with the goal of enabling practical use without sacrificing fidelity.}}

\neweacl{Automatically generating LFs requires of techniques capable of producing \reveswa{a formal representation from text,} following a set of pre-defined grammar rules. This challenge is commonly addressed in so-called semantic parsing tasks \citep{yin17acl, Radhakrishnan-2020}, \reveswa{but they have not been applied to table-to-text before}. For instance, \citet{Guo2019} present IRNet, a NL-to-SQL semantic parser that generates grammatically correct SQL sentences based on their natural language descriptions. Valuenet, introduced by \citet{Brunner2021}, presents a BERT-based encoder \citep{Devlin2018} in IRNet. In this work, we adapted the grammar-based decoder of Valuenet to produce LFs, which allowed us to show that we can produce high quality LFs.}

\section{Model}
\label{sec:model}

\reveswa{In this section we first introduce Logical Forms, and then the model that produces descriptions for tables via automatically produced Logical Forms.}

\subsection{Logical Forms}
\label{sec:lf}
The LFs used in this work are tree-structured logical representations of the semantics of a table-related statement, similar to AMR graphs \citep{Banarescu2012}, and follow the grammar rules defined by \citep{Chen2020a}. Each rule can be executed against a database, a table in this case, yielding a result based on the operation it represents. As these graphs represent factual statements, the root is a boolean operation that should return \reveswa{True uppon execution}. 
Figure \ref{fig:lf-example} shows an example of a table with its caption and logical form.

\subsubsection{Logical Form Grammar}
\label{subsec:lf-grammar-mod}

\begin{figure}[t]
 \begin{minipage}{0.48\linewidth}
  \small
  \flushleft
 \textbf{Caption:}
 
 1979 philadelphia eagles season
\vspace{0.1cm}

 \textbf{Table:}
 \vspace{0.1cm}
 
  \def\arraystretch{1.2}
\setlength\tabcolsep{2.0pt}
  \begin{tabular}{|c|c|r|}
\hline 
\rowcolor[HTML]{ECECEC} 
opponent            & result    & attendance \\ \hline 
new york giants     & w 23-17 & 67000      \\ \hline
atlanta falcons     & l 14-10 & 39700      \\ \hline
new orleans saints  & w 26-14 & 54000      \\ \hline
new york giants     & w 17-13 & 27500      \\ \hline
pittsburgh steelers & w 17-14 & 61500      \\ \hline
\end{tabular}
\vspace{0.1cm}

\textbf{Statement:} In the 1979 Philadelphia Eagles season there was an average attendance of 52500 in all winning games.
 \end{minipage}\hfill
 \begin{minipage}{0.48\linewidth}
 \small
 \flushleft
  \textbf{LF:} 
eq \{ avg \{ filter\_str\_eq \{ all\_rows ; result ; w \} ; attendance \} ; 52500 \} = \reveswa{True}
  \includegraphics[width=\linewidth]{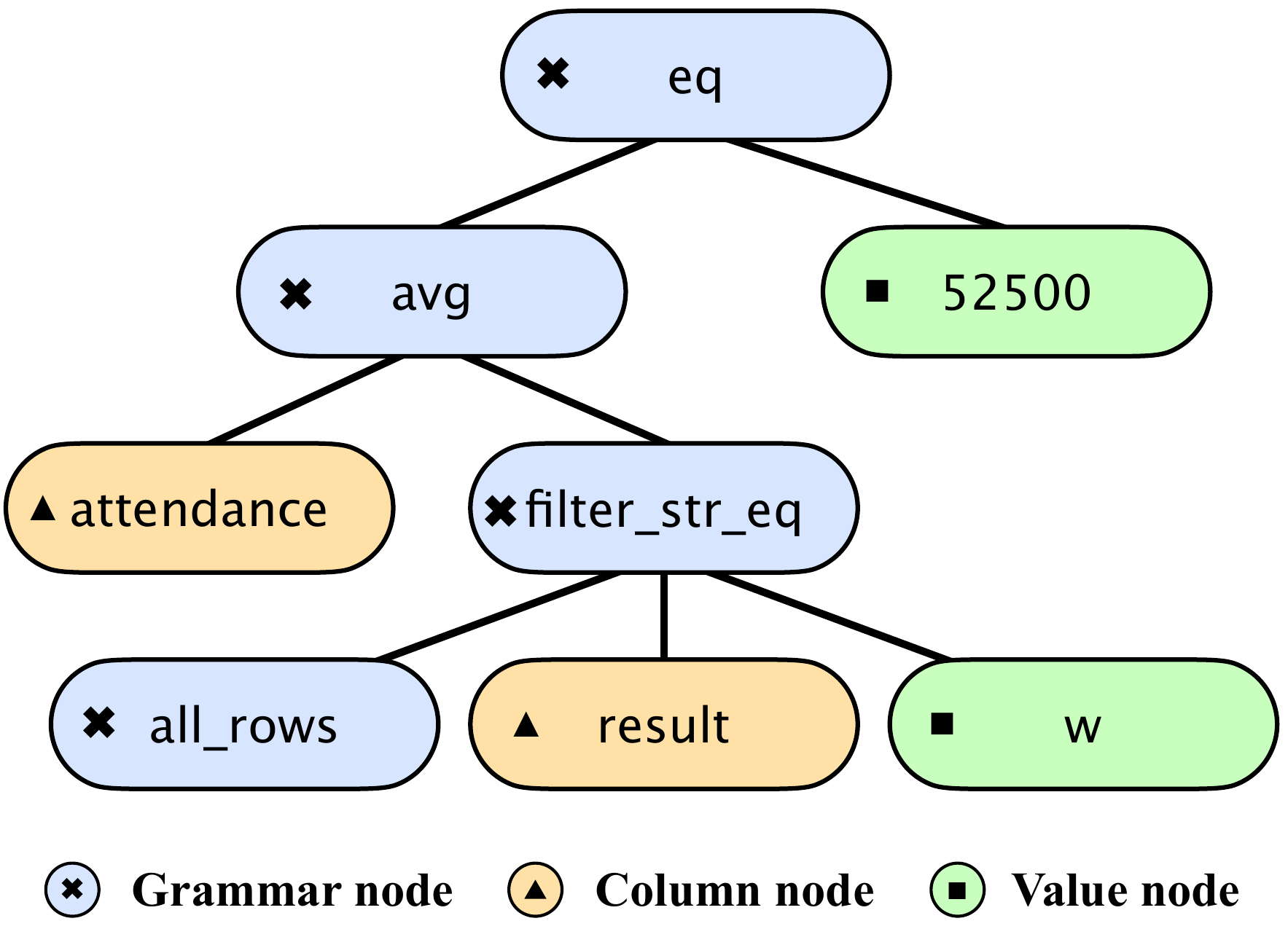}
  \small
\textbf{Content Selection values:} 52500, w 
 \end{minipage}
 \caption{Example of a table with its caption, a logical form (in linearized and graph forms)\neweacl{, its corresponding content selection values} and the target statement. Note that \emph{w} in the table stands for \emph{win}. 
More details in the text. 
}
 \label{fig:lf-example}
\end{figure}

The grammar contains several non-terminals (nodes in the graph, some of which are illustrated in Fig. \ref{fig:lf-example}), as follows: 

\neweacl{
\textbf{Stat} represents boolean comparative statements such as greater than, less than, equals (shown as \emph{eq} in the figure), not equals, most equals or all equals, among others. This is the root of the LF graph. 

\textbf{C} refers to an specific column in the input table (\emph{attendance} and \emph{result} in the figure).

\textbf{V} is used for specific values, which can be either values explicitly stated in the table (\emph{w} in the figure) or arbitrary values used in comparisons or filters (\emph{52500} in the figure).}

\neweacl{\textbf{View} refers to a set of rows, which are selected according to a filter over all rows. The filters refer to specific conditions for the values in a specific column, e.g. \emph{greater}. 
The figure shows \emph{all\_rows}, which returns all rows, and also \emph{filter\_str\_eq} which returns the rows that contain the substring \emph{``w"} in the \emph{result} column.}

\neweacl{\textbf{N} is used for operations that return a numeric value given a view and column as input, such as sums, averages (shown as \emph{avg} in the figure), maximum or minimum values, and also counters.}

\neweacl{\textbf{Row} is used to select a single row according to maximum or minimum values in a column.}

\neweacl{\textbf{Obj} is used for operations that extract values in columns from rows (either views or specific rows). The most common operations are \emph{hop} extractors that extract a unique value, for instance \emph{str\_hop\_first} extracts a string from the first row of a given \emph{View}.}

\neweacl{\textbf{I} is used to select values from ordinal enumerations in \emph{N} and \emph{Row} rules, as for instance in order to select the ``the 2nd highest" \emph{I} would equal to 2.}

Please refer to the \ref{app:lf-grammar} for full details. \reveswa{Keep in mind that} \emph{Stat}, \emph{View}, \emph{N}, \emph{Row} and \emph{Obj} are internal nodes that constitute the structure of the LF (shown in blue in the figure), while column \emph{C}, value \emph{V} and index \emph{I} nodes are always leaf nodes. 

We \reveswa{identified} several ambiguities in the original grammar formulation that \reveswa{hindered the training of a semantic parser producing LFs.}

\neweacl{The first one affects all functions that involve strings. Within the LF execution engine proposed by \citet{Chen2020a}, the implementation of those functions are divided into two: one that handles numeric and date-like strings, and a strict version for other string values. \reveswa{As a result, we explicitly represented these as two distinct functions within the grammar: a group for numerical and date-like values, and an additional group for other string values, denoted by the suffix "\_str". }}
\reveswa{The second issue addresses an inconsistency with the} \emph{hop} function. This function, when provided with a \emph{Row}, returns the value associated to one of its columns. Although the grammar specifies that these functions are exclusively applied to \emph{Row} objects, in 25\% of the dataset examples, the function is used on a \emph{View} object instead, which can encompass multiple rows. To address this, we defined a new function \emph{hop\_first} tailored to these specific situations.

The grammar in \ref{app:lf-grammar} contains the new rules that fix the ambiguity issues.
We also converted automatically each LF in the dataset to conform to the unambiguous grammar. The conversion script is publicly available.

\subsubsection{Content Selection}
\label{subsec:extra-values}

\neweacl{\reveswa{To isolate the impact of content selection and full LFs, we extracted the LF values, allowing us to evaluate model performance with and without content selection. These extracted values include those explicitly stated in table cells, as well as other values existing in the LF but not explicitly present in the table, such as results of arithmetic operations.} This set of values constitutes the supplementary input to the systems when using content selection (CS for short), categorized as follows:}

\begin{itemize}
\item 
\textbf{TAB}: Values present in a table cell, verbatim or as a substring of the cell values. 

Figure \ref{fig:lf-example} shows an example, where ``w" is a substring in several cells. 72.2\% of the values are of this type.

\item 
\textbf{INF}: Values not in the table that are inferred, e.g. as a result of an arithmetic operation over values in the table. For instance 52500 in Figure \ref{fig:lf-example} corresponds to the average over attendance values. 
20.8\% of \emph{Value} nodes are INF.

\item 
\textbf{AUX}: Auxiliary values not in the table nor INF that are used in operations, e.g. to be compared to actual values in cells, as in ``\emph{All scores are bigger than 5.}". Only 7.1\% are of type AUX. 
\end{itemize}

\neweacl{In principle, one could train a separate model to select and generate all necessary content selection values for input into any Table-to-Text model, as follows: 1) Choose values from table cells, whether in full or as substrings (TAB); 2) Infer values through operations like average, count, or max (INF); 3) Induce values for use in comparisons (AUX). In order to separate the contribution of content selection and the generation of LFs, we chose to focus on using content selection and not yet on producing the actual values. Hence, we derive these values from the manual gold \reveswa{reference} LFs, \reveswa{i.e., human-made reference logical forms provided in the dataset}, and feed them to the models. The experiments will demonstrate that this content selection step is critical, and that current models fail without it. We leave the task of automatic content selection for further research.}

\begin{figure}[t!]
\includegraphics[width=\linewidth]{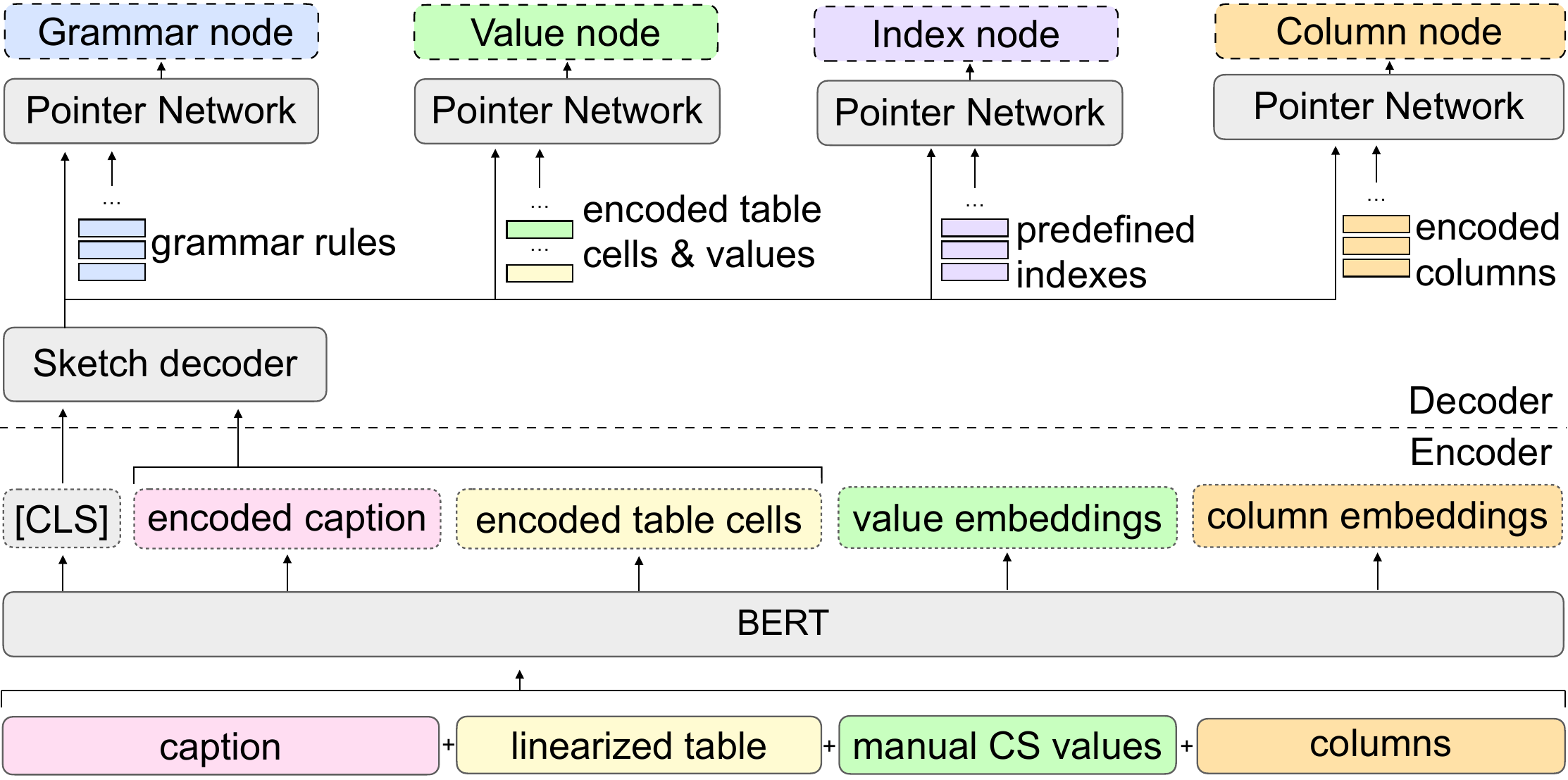}
\caption{Table2Logic architecture, with input in the top and output in the bottom. See text for details. 
}
\label{fig:t2l-architecture}
\end{figure}

\subsection{Generating Text via Logical Forms}
\label{sec:system}

Our Text-to-Logic-to-Text (\tlt) system has two main modules in a pipeline: 

Given a table, its caption and, optionally, selected content, \neweacl{\textbf{Table2Logic} generates an LF; With the same table information, plus the generated LF, \textbf{Logic2Text} produces the statement text. 
}

\subsubsection{Table2Logic Module}
\label{subsec:grammar-based-decoder}

\neweacl{We frame this model as semantic parsing, adapting the \neweacl{IRNet grammar-based decoder by \citep{Guo2019}} to LFs. More specifically, we follow the implementation of Valuenet by \citet{Brunner2021}, which is a more up to date revision of IRNet. Both models are NL-to-SQL semantic parsers that generate grammatically correct SQL sentences based on their descriptions. We adapted the system to produce logical forms instead of SQL. }\neweacl{The architecture of Table2Logic is presented in Figure \ref{fig:t2l-architecture}}. 

We first feed a pre-trained BERT encoder \citep{Devlin2018} with the concatenation of the following table data: 
the caption text, the table content in linearized form, the column names, and, in some of our model configurations, a set of content selection values manually extracted from the associated gold reference LF. \reveswa{The details about content selection values are presented in Section \ref{subsec:extra-values}.} 

The output embeddings of the \emph{CLS} token, the caption tokens and the linearized values in the table are fed into an LSTM decoder \citep{Hochreiter1997}. At each decoding step, the attention vector of the LSTM is used by four different pointer networks \citep{Vinyals2015}. \reveswa{Each of these pointer networks specializes in generating one node type: \emph{grammar}, \emph{Value}, \emph{Column} and \emph{Index}. We follow a constrained decoding strategy where a pointer network is selected based on the node type that should follow the previously generated ones according to the grammar of LFs. Each of these pointer networks utilize the previously mentioned attention vector alongside a set of embeddings. In the case of \emph{Value} and \emph{Column} node types, these embeddings consist of the CS values and column encodings produced by the BERT model. On the other hand, \emph{Index} and \emph{grammar} node types use a separate set of predefined embeddings associated to each ordinal index and LF grammar rule respectively.}

\neweacl{Following \citep{Guo2019}, Table2Logic performs two decoding iterations. In a first iteration, a sketch LF is generated using the grammar pointer network. The sketch LF consisting only of grammar related nodes (e.g. those in blue in Fig. \ref{fig:lf-example}), where \emph{Value}, \emph{Column} and \emph{Index} nodes are represented by placeholders that are filled in a second decoding iteration by the corresponding pointer network. }

\neweacl{We follow a teacher-based training strategy to calculate the loss for each decoding iteration. In the first iteration the loss is calculated by accumulating the cross entropy loss for each generated grammar node given the previous gold reference nodes. The sketch is then used to calculate the cross entropy loss of generating \emph{Value}, \emph{Column} and \emph{Index} nodes. The weights of the network are updated using the sum of both loss values.}

During inference, we use beam search to produce a set of candidates. In addition, we explore a \neweacl{False Candidate Rejection (FCR)} policy to filter out all LFs in the beam representing a \emph{False} statement, as they would lead to a factually incorrect sentence. \reveswa{As previously mentioned in \ref{sec:lf}, the root node of each LF always consists of a boolean grammar rule. The structured nature of LFs enables us to automatically execute them against a data source, in this case, the table. Consequently, each LF yields either \emph{True} or \emph{False} based on the relationships between the various facts it encompasses. We exploit this property of LFs to discard all generated LFs that, despite their grammatical correctness, convey a \emph{False} statement.} Thus, only the candidate LF in the beam that executes to \emph{True} with maximum beam probability is be selected. Section \ref{sec:dev-experiments} reports experiments with FCR.

\subsubsection{Logic2Text Module}
\label{subsec:logic2text}
For the language realization model we use the top performer in \citep{Chen2020a}. \reveswa{This model consists on a GPT-2 \cite{Radford2019} pre-trained large language model (LLM) fine-tuned to generate text from tables and human-produced logical forms. The implementation is rather simple; the input sequence is a concatenation of the table caption, table headers, and the linearized table content and logical form. The model, referred to as Logic2Text, receives this input and generates a sentence that is strongly conditioned by the semantic represented by the provided LF. The Logic2Text model enables us to produce natural language statements based on the automatic LFs produced by our Table2Logic model.}

\section{Experiments}
\label{sec:experiments}
\neweacl{In this section we report the results on text generation using the test split of the Logic2Text dataset. We first introduce the dataset, the different models, the automatic evaluation and the manual evaluation.}

\subsection{Dataset}
\label{subsec:dataset}

\neweacl{We use the dataset introduced by \citet{Chen2020a}, \reveswa{a human-annotated dataset comprising 4992 open-domain tables obtained from the LogicNLG dataset \citep{Chen2020b}. Each table is paired with an average of 2 human-written statements describing facts within the table. Following a predefined questionnaire, each annotator describes the logic behind these statements. Subsequently, \citet{Chen2020a} use the given answers to derive the LFs associated with each statement. The resulting dataset contains a total of 10753 examples (8566 train, 1092 dev. and 1095 test) of high quality human-produced LFs alongside its corresponding statement and table information. We refert to these manually produced LFs as gold LFs, in contrast to the automatic LFs produced by our model. As mentioned in the introduction, Table-to-Text tasks are underspecified, allowing many other statements (and LFs) not provided in the dataset to be factually correct and equally informative as the ones in it.}} 

\subsection{Model Configurations}
\label{sec:model-config}
\begin{figure*}[t]
\includegraphics[width=\linewidth]{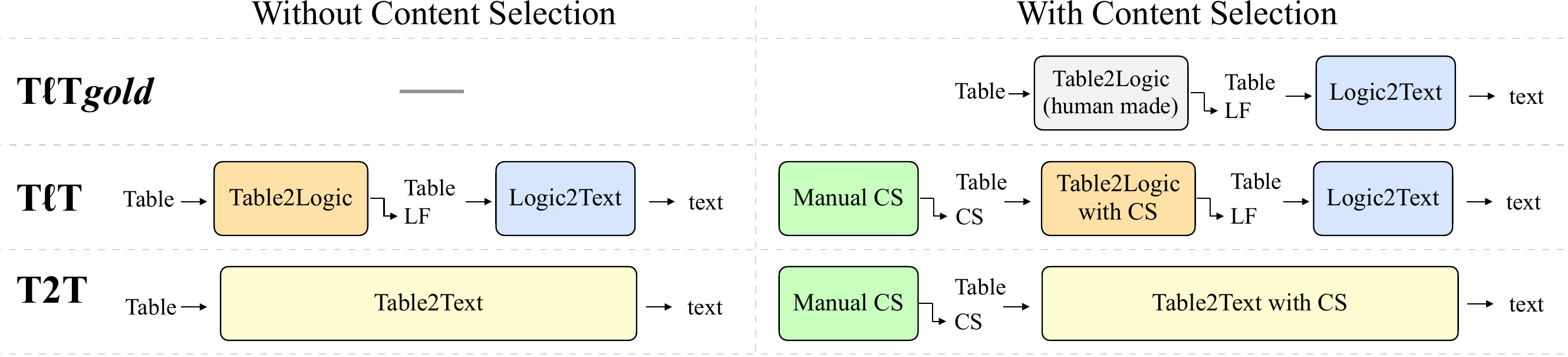}
\caption{Model configurations used in the main experiments.}
\label{fig:models}
\end{figure*}

The configuration of the different models are shown in Figure \ref{fig:models}. All models take as input the table information, including table caption, linearized table and column headers. In the top row, we include the upperbound system \tlt$_{gold}$, which takes the table plus the manually produced gold reference LF as input. In the middle row we include our system \tlt, which is composed by the Table2Logic module and the Logic2Text module. Both \tlt~ and \tlt$_{gold}$ use the same Logic2Text module, but while the first uses automatically produced LFs, the second uses manual LFs. \tlt~ is evaluated in two variants, with and without content selection (\tlt~ and \tlt$_{noCS}$, respectively). Logic2Text uses default hyperparameters \citep{Chen2020a}.

The bottom row shows our baselines (T2T, short for Table2Text), which generate the text directly from table information, with and without content selection data. \reveswa{Since Logic2Text is based on state-of-the-art generation \citep{Chen2020a}, and to ensure compatibility, both T2T and T2T$_{noCS}$ have the share codebase.} \neweswa{That is, T2T uses the same GPT-2 model architecture as in \cite{Chen2020a} but trained without LFs. Receiving only the linearized table (in case of T2T$_{noCS}$) and, in the case of T2T, the same list of manual CS values as \tlt.}

\subsection{\reveswa{Content Selection Ablation Study}}
\label{sec:dev-experiments}
In order to develop Table2Logic, \reveswa{we examined the influence of content selection, along with the impact} of rejecting LFs that evaluate to \emph{False} (FCR) in development data. Accuracy was computed using strict equality with respect to any of the manual gold reference LFs. Both sketch accuracy (using placeholders for non-grammar nodes) and full accuracy are reported. As mentioned in the introduction, this task is underspecified, in that multiple LFs which are very different from the gold reference LFs could be also correct. Still, the accuracy is a good proxy of quality to discriminate between better and worse models. The results correspond to the checkpoints, out of 50 epochs, with the best full accuracy on development.  We tuned some hyperparameters on development and used default values for the rest (see \ref{app:hyperparameters} for details).

Table \ref{tab:sketch-full-acc} shows the results for different subsets of content selection values, with the last row reporting results when FCR is used. Without FCR, the most important set of values are those explicit in the table (TAB), and the best results correspond to the use of all values, although AUX values do not seem to help much (in fact, the best non-FCR full results are obtained without using AUX, by a very small margin). 
The last row reports a sizeable improvement in accuracy for full LFs when using FCR, showing that FCR is useful to reject faulty LFs that do not evaluate to True.

Overall, the full accuracy of \tlt ~might seem low, but given that the gold reference LFs only cover a fraction of possible LFs they are actually of good quality, as we will see in the next sections.

 \neweacl{We also performed an additional ablation experiment where we removed the table information from the system in the last row (\tlt). 
 The sketch and full accuracies dropped ($50.3$ and $42.7$ respectively), showing that access to table information is useful even when content selection is available.}

\begin{table}[]
\setlength\tabcolsep{3.0pt}
\begin{center}
\begin{tabular}{lrr}
\hline
Model              & Sketch    & Full      \\ \hline
No content selection (\tlt$_{noCS}$)         & 15.0          &  4.9          \\ \hline
AUX                & 14.0          &  6.2          \\
INF                & 28.7          & 11.0          \\
TAB                & 42.6          & 27.3          \\
TAB, INF           & 56.5          & 39.3          \\
TAB, AUX           & 44.3          & 28.6          \\ 
TAB, INF, AUX      & \textbf{58.5} & 38.9          \\ \hline
TAB, INF, AUX + FCR (\tlt) & 56.0          & \textbf{46.5} \\ \hline
\end{tabular}
\end{center}
\caption{\neweacl{Table2Logic: Accuracy (\% on dev.) over sketch and full versions of gold LFs using different subsets of content selection (CS) and FCR in development. First row for \tlt$_{noCS}$, last row for \tlt, as introduced in Sect. \ref{sec:experiments}.}}

\label{tab:sketch-full-acc}
\end{table}

\subsection{Automatic Evaluation}
\label{subsec:automatic-eval}
The automatic metrics compare the produced description with the reference descriptions in the test split. As shown in Table \ref{tab:auto-metrics}, we report the same \reveswa{n-gram similarity} automatic metrics as in \citep{Chen2020a}, BLEU-4 (B-4) \citep{papineni-etal-2002-bleu}, ROUGE-1, 2, and L (R-1, R-2, and R-L for short) \citep{lin-2004-rouge}, \neweswa{along with two additional metrics BERTscore (BERTs) \citep{zhang2019bertscore} and BARTscore (BARTs) \citep{yuan2021bartscore} which can capture the semantic similarity between the ground truth and generation results}. The results show that generation without content selection is poor for both the baseline system and our system (T2T$_{noCS}$ and \tlt$_{noCS}$, respectively). Content selection is key for good results in both kinds of systems, which improve around 10 points in all metrics when incorporating content selection (T2T and \tlt). Automatic generation of LFs (\tlt) allows to improve over the system not using them (T2T) in at least one point. If \tlt~ had access to correct LFs it would improve 4 points further, as shown by the \tlt$_{gold}$ results. \reveswa{Observe} that our results for \tlt$_{gold}$ are very similar to those reported in \citep{Chen2020a}, as shown in the last row. We attribute the difference to minor variations in the model released by the authors. 

\begin{table}[]
\setlength\tabcolsep{2.0pt}
\begin{center}
\begin{tabular}{lcccccc}
\hline
Model      & B-4            & R-1            & R-2            & R-L       & \neweswa{BERTs} & \neweswa{BARTs}    \\ \hline
T2T$_{noCS}$      & 16.8          & 37.7          & 19.3          & 31.6   & \neweswa{88.8} & \neweswa{-4.04}       \\
\tlt$_{noCS}$  & 15.6          & 39.0          & 18.9          & 32.2   & \neweswa{87.9} & \neweswa{-4.03}        \\ \hline
T2T     & 26.8          & 55.2          & 31.5          & 45.7  & \neweswa{91.9} & \neweswa{\textbf{-2.98}}       \\
\tlt ~(ours) & \textbf{27.2} & \textbf{56.0} & \textbf{33.1} & \textbf{47.7} & \neweswa{\textbf{92.0}} & \neweswa{-2.99} \\ \hline
\tlt$_{gold}$    & 31.7          & 62.4          & 38.7          & 52.8   & \neweswa{93.1} & \neweswa{-2.65}       \\ 
\tlt$_{gold}$*   & 31.4*          & 64.2*          & 39.5{*}          & 54.0{*}  & \neweswa{-} & \neweswa{-}        \\ \hline
\end{tabular}
\end{center}
\caption{Automated \reveswa{n-gram similarity} metrics for textual descriptions (test). \reveswa{BLEU-4 (B-4), ROUGE-1, 2, and L (R-1, R-2, and R-L), BERTscore (BERTs) and BARTscore (BARTs)}. Bottom two rows are upperbounds, as they use manual LFs. See text for system description. * for results reported in \cite{Chen2020a}. \neweswa{Both BERTs and BARTs correspond to the f1 score. In case of the BARTscore higher is better.}
}
\label{tab:auto-metrics}
\end{table}

\subsection{\neweswa{Human} Fidelity Evaluation}
\label{subsec:fidelity-eval}
\neweacl{Given the cost of \neweswa{human} evaluation, we selected three models to manually judge the fidelity of the produced descriptions: the baseline T2T model, our \tlt~ model and the upperbound with manual LFs, \tlt$_{gold}$. For this,} we randomly selected 90 tables from the test set and generated a statement with each of the three models. In order to have two human judgements per example, \neweswa{we provided each evaluator with 30 sentences, along with the corresponding table and caption.} The evaluators were asked to select whether the description is true, false or nonsense according to the caption and the table.\neweswa{This group of evaluators was comprised of eighteen volunteer researchers unrelated to this project.} \reveswa{We use Fleiss' kappa coefficient \citep{fleiss1971measuring} to measure the inter-evaluator agreement. This coefficient is a statistical measure used to assess the level of agreement among multiple raters when categorizing items into different classes. It takes into account both the observed agreement and the agreement expected by chance. It is a way to determine the extent to which the agreement among raters goes beyond what would be expected due to random chance alone. The coefficient ranges from -1 to 1, where higher values indicate better agreement beyond chance, while lower values indicate poor agreement.} \neweswa{The evaluation concluded with a strong 0.84 Fleiss' kappa coefficient}. We discarded examples where there was disagreement.

Table \neweacl{\ref{tab:fidelity-eval-results}} shows the fidelity figures for the three models. After the evaluation, we noticed that the faithfulness results for \tlt$_{gold}$ in our experiment matched the figure reported by \citet{Chen2020a}, so we decided, for completeness, to include in the table their figures for T2T$_{noCS}$, which should be roughly comparable to the other results in the table.

In general, the differences in human fidelity evaluation are much higher than for automatic metrics, which we attribute to widely recognised issues of automatic metrics when evaluating text generation. \reveswa{In our case, the two most significant issues are the ones affecting n-gram overlaping metrics (e.g., BLUE, ROUGE). These automatic metrics exhibit insensitivity to semantic and pragmatic quality, making them fail to capture the semantic and pragmatic nuances of language. This can lead to models generating text that, despite being technically correct in terms of word overlap, can still be semantically inaccurate \citep{zhang2019bertscore}. Furthermore, these metrics can also suffer from a lack of correlation with human judgment, leading to models that could generate text that is grammatically correct but incoherent and meaningfulness, yet receives a high score \citep{moramarco-etal-2022-human}.} From low to high, the results allow us to estimate the \textbf{separate contributions} of each component in absolute fidelity points: 
\begin{itemize}
\item \textbf{Manual content selection} improves fidelity in 24 points (T2T$_{noCS}$ 
vs. T2T) ; 
\item \textbf{Automatic LFs} improve an additional 30 points (T2T vs. \tlt); 
\item \textbf{Manual LFs} give 7 points (\tlt ~vs. \tlt$_{gold}$); 
\item \textbf{Perfect Logic2Text} generation would yield 18 points (\tlt$_{gold}$ vs. 100\%). 
\end{itemize}

The figures confirm our contribution: it is possible to produce logical forms automatically, and they allow to greatly improve fidelity, with the largest fidelity improvement in the table, 30 absolute points, \reveswa{which correspond to a 67\% improvement over the comparable system not using LFs}. Note that the other improvements are actually gaps which allow us to prioritize the areas for further research: automatic content selection (24 pt.), better Logic2Text (18 pt.) and better Table2Logic (7 pt.). In the following section we analyse the errors in the two later modules.

\begin{table}[]
\centering
\begin{tabular}{lccc}
\toprule
Model      & Faithful      & Unfaithful    & Nonsense     \\ \midrule
T2T$_{noCS}$*     & 20.2*          & 79.8*          & - \\
T2T     & 44.9          & 49.3          & 5.8 \\
\tlt~ (ours) & \textbf{75.0} & \textbf{20.3} & \textbf{4.7}          \\ \midrule
\tlt$_{gold}$    & 82.4          & 13.51          & 4.1          \\ \bottomrule
\end{tabular}
\caption{\reveswa{Human evaluation fidelity results. Given 90 test samples to three different model configurations, percentage of generated sentences identified as Faithful, Unfaithful or Nonsense by evaluators. Answer with full disagreement between evaluators are discarded. }
* for results reported in \citep{Chen2020a}.
}
\label{tab:fidelity-eval-results}
\end{table}

\subsection{Qualitative Analysis}
\label{sec:qualitative}
\neweacl{We performed a qualitative analysis of failure cases in both Table2Logic and Logic2Text, as well as examples of factually correct descriptions generated from LFs different from gold LFs. }

\subsubsection{Table2Logic}
\label{subsec:qualitative-table-to-logic}

\begin{table}[t]
\fontsize{10}{10}\selectfont
\def\arraystretch{1.5}
\centering
\begin{tabular}{lllll}\hline
\textbf{} & \textbf{Fr.} & \textbf{Total}          & \textbf{Confusions}                            \\
\hline
Stat                          & 0.38                             & 0.13                       & \makecell{greater → less \\ all equals → most equals \\ equals → and}                                \\
\hline
C                             & 0.25                             & 0.19                       & \makecell{column 3 → column 0 \\ column 1 → column 0}                                              \\
\hline
Row                           & 0.16                             & 0.02                       & \makecell{row 0 → row 2 \\ row 2 → row 0 \\ row 2 → row 1}                                           \\
\hline
View                          & 0.11                             & 0.20                       & \makecell{filter\_greater → filter\_less \\ filter\_greater → filter\_eq \\ filter\_eq → all\_rows}  \\
\hline
N                             & 0.05                             & 0.03                       & \makecell{sum → avg \\ avg → sum}\\
\hline
Obj                           & 0.03                             & 0.26                       & \makecell{str\_hop → num\_hop \\ num\_hop → str\_hop}          \\
\hline
V                             & 0.01                             & 0.16                       & \makecell{value 72 → value 73 \\ value 70 → value 71} \\
\hline
I                             & 0.01                             & 0.01                       & 1 → 0 \\
\hline
\end{tabular}
\caption{Table2Logic: Distribution of differing node types (\tlt~ vs. gold LFs). Fr. for frequency of node type in differing LFs, Total for overall frequency in gold. Rightmost column for most frequent confusions (\tlt~  → gold).
}
\label{tab:qualitative_dev}
\end{table}

\neweacl{
We automatically compared the LFs generated by \tlt~ in the development set that did not match their corresponding gold LFs. Note that the produced LFs can be correct even if they do not match the gold LF. We traverse the LF from left to right and record the first node that is different. Table \ref{tab:qualitative_dev} shows, in decreasing order of frequency, each grammar node type  (cf. Section \ref{subsec:lf-grammar-mod}) with the most frequent confusions. }

\neweacl{
The most frequent differences focus on \emph{Stat} nodes, where a different comparison is often generated. The next two frequent nodes are column and row selections, where \tlt~ selects different columns and rows, even if \tlt~ has access to the values in the content selection. The frequency of differences of these three node types is well above the distribution in gold LFs. The rest of differences are less frequent, and also focus on generating different comparison or arithmetic operations.}

\begin{table*}[t]
\fontsize{10}{10}\selectfont
\def\arraystretch{1.5}
\setlength{\tabcolsep}{3pt}
\centering
\begin{tabular}{p{3cm}p{10cm}}
\textbf{LF difference} & \textbf{Sentences}                            \\
\hline
Similar structure, semantically equivalent
&
\begin{tabular}[t]{p{10cm}}
\textbf{\tlt}: In the list of Appalachian regional commission counties, Schoharie has the highest unemployment rate.\\\textbf{Human}: The appalachian county that has the highest unemployment rate is Schoharie.
\end{tabular} \\
\hline
Similar structure, semantically different                          & 
\begin{tabular}[t]{p{10cm}}
\textbf{\tlt}: Dick Rathmann had a lower rank in 1956 than he did in 1959.\\\textbf{Human}: Dick Rathmann completed more laps in the Indianapolis 500 in 1956 than in 1959.  
\end{tabular}
\\
\hline
Different structure, semantically different                          & 
\begin{tabular}[t]{p{10cm}}
\textbf{\tlt}: Most of the games of the 2005 Houston Astros' season were played in the location of arlington.\\\textbf{Human}: Arlington was the first location used in the 2005 Houston Astros season.  
\end{tabular}
\\
\hline
Simpler structure, more informative                          & 
\begin{tabular}[t]{p{10cm}}
\textbf{\tlt}: Aus won 7 events in the 2006 asp world tour.\\\textbf{Human}: Seven of the individuals that were the runner up were from aus.  
\end{tabular}
\\
\end{tabular}
\caption{Examples of faithful sentences produced by \tlt~ from intermediate LFs that do not match the gold LF. 
}
\label{tab:qualitative_no-gold-faithful}
\end{table*}

\subsubsection{Logic2Text}
\label{subsec:qualitative-logic-to-text}
\neweacl{The faithfulness score of descriptions generated from gold LFs (\tlt$_gold$) is 82\%, so we analysed a sample of the examples in this 18\%. For the sake of space, we include full examples in Appendix \ref{app:qualitative-logic-to-text-examples}, which include table, caption, gold LF and generated description. We summarize the errors in three types: }

\neweacl{
\noindent \textbf{Comparative arithmetic}: Logic2Text miss-represented comparative arithmetic action rules in the LF in 40\% of the cases. This resulted in cases where the output sentence declared that a given value was \textit{smaller} than another when the LF stated it was \textit{larger}. Logic2Text also seem to ignore \textit{round} and \textit{most} modifiers of comparison operations, producing sentences with strict equality and omitting qualifiers like ``roughly" or ``most". The absence of these qualifiers made the produced sentences factually incorrect. 

\reveswa{The reason behind these types of errors remain uncertain. One plausible explanation could be linked to the limited number of parameters within the models of this architecture. While these models are capable of recognizing the need for a comparative rule at a given step, their size may still be insufficient for effectively distinguishing between two potential comparisons of the same category, e.g. \textit{smaller} and \textit{larger}. Another contributing factor may be related to the small amount of occurrences of each type of comparative rule within the training dataset. Only 44\% of LFs in the training set contain any of the 22 comparative arithmetic action rules. Finally, we must also highlight that models that do not use LFs also incur in these kind of errors, showing that these are common errors across different model architectures and are not exclusive to our specific model.} 
}

\neweacl{\noindent \textbf{LF omission}: Logic2Text disregarded part of the LF (33\% of errors), resulting in omissions that led to false sentences. Many of these errors involved omitting an entire branch of the LF, leading, for instance, to sentences wrongly referring to all the instances in the data instead of the subset described in the LF.}

\neweacl{\noindent \textbf{Verbalization}: Logic2Text incurred in wrong verbalization and misspellings (27\% of cases). For instance Logic2Text producing a similar but not identical name like in \emph{foulisco} instead of \emph{francisco}. }

\neweacl{We attribute the errors to the fact that the generator is based on a general Language Model such as GPT-2. While these language models are excellent in producing fluent text, it seems that, even after fine-tuning, they have a tendency to produce sentences that do not fully reflect the data in the input logical form. It also seems that the errors might be explained by the lower frequency of some operations. The 18\% gap, even if it is much lower than the gap for systems that do not use LFs, together with this analysis, show that there is still room for improvement.
}

\subsubsection{\reveswa{Implications of Divergent LF Production from Gold Reference LF}}
\label{subsec:qualitative-no-gold-faithful}
\neweacl{The results in Table \ref{tab:sketch-full-acc} show that our Table2Logic system has low accuracy when evaluated against gold logical forms (46\%). On the contrary, the results in fidelity for the text generated using those automatically generated logical forms is very high, 75\%, only 7 absolute points lower to the results when using gold logical forms. This high performance in fidelity for automatic LFs might seem counter-intuitive, but we need to note that it is possible to generate a correct and faithful LF that is completely different from the gold logical form, i.e. the system decides to produce a correct LF that focuses on a different aspect of the information in the table with respect to the gold LF. }

\neweacl{In order to check whether this is actually the case, we manually examined the automatic LFs from \tlt~ that resulted in faithful sentences in the manual evaluation while being ``erroneous", that is, different from their gold LF references.  
In all cases, such \tlt~ LFs are correctly formed and faithful, i.e.  even if these LFs where ``wrong" according to the strict definition of accuracy, the semantics they represent are informative and faithful to the source data. Table \ref{tab:qualitative_no-gold-faithful} shows a sample of the output sentence, with full details including table and LFs in \ref{app:faithful-diff-lf-full-examples}. 

We categorized the samples as follows. 69\% of them share a similar LF structure as their corresponding gold references, but with changes in key \emph{Value} or \emph{Column} nodes, making them semantically different. In 15\% of the cases the LF had similar structure, but although there were differences, the LF was semantically equivalent to the gold LF. The rest of \tlt~ LFs (16\%) had a different structure, and where semantically different from reference counterparts, while still being correct and faithful to the table. \reveswa{This reflects an interesting aspect of reference-based evaluation. In many cases, generating a sentence that diverges from the reference does not imply that such a sentence is less faithful, useful or informative. Thus, the accuracy evaluation with respect to gold LFs (cf. Table \ref{tab:sketch-full-acc}) provides an underestimate of the quality of the produced LFs and texts.}

All in all the quality of LFs and corresponding text produced by \tlt~ for this sample is comparable to those of the gold LF, and in some cases more concise and informative. }
\neweacl{This analysis confirms that the quality of Table2Logic is well over the 46\% accuracy estimate, and that it can be improved, as the produced text lags 7 points behind gold LFs.}

\section{Conclusions and \reveswa{F}uture \reveswa{W}ork}
\label{sec:conclusions-fw}

We have presented \tlt ~which, given a table and a selection of the content, first produces logical forms and then the textual statement. We show for the first time that automatic LFs improve results according to automatic metrics and, especially, manually estimated factual correctness. In addition, we separately study the contribution of content selection and the formalization of the output as an LF, showing a higher impact in fidelity of the later. In this paper, our focus is on tables. However, our findings and software can readily be extended to other structured inputs. \reveswa{Given that the grammar of LFs is independent of the table format, it can be easily adjusted for other common data-to-text inputs such as graphs or triplets by modifying its execution engine, keeping the LFs intact}.

\reveswa{Our contribution enables future Data-to-Text applications to leverage the advantages of using factually verifiable logical forms, eliminating the need of manually constructed LFs. These advantages include a relative improvement in fidelity of 67\% compared to baseline models, along with the ability to access an intermediate formal representation within the generation process. This facilitates the automated validation of a statement's factual accuracy before generating its corresponding natural language representation. The improvement in fidelity attained by our model is relevant for most Data-to-Text applications, where faithfulness is crucial.}

\reveswa{The conducted analysis also} enabled us to quantify that content selection would offer the most substantial performance improvement, followed to a lesser extent by improved logic-to-text generation, and, finally, improved table-to-logic generation. In the future, we plan to focus on automatic content selection, which we think can be largely learned from user preference patterns found in the training data. \reveswa{Recent advances in semantic parsing, e.g. the use of larger language models \citep{Raffel2019, bigscience_workshop_2022, Zhang2022}, could also be easily folded in our system and would further increase the contribution of LFs.} Finally, we also plan \neweacl{to make use of our qualitative analysis} to explore complementary approaches for improving factual correctness in logic-to-text.

\section*{Acknowledgements}

This work is partially funded by MCIN/AEI 10.13039/501100011033 and by the European Union NextGenerationEU/ PRTR, as well as the Basque Government IT1570-22.

\bibliographystyle{elsarticle-harv} 
\bibliography{bibliography}

\appendix

\section{Training Procedure}
\label{app:reproducibility}
All experiments where carried out in a machine with a GPU NVIDIA TITAN Xp 12GB. The average training runtime for all Table2Logic based models is 19 hours. For the Logic2Text presented models, it averaged 10 hours. Both Table2Logic and Logic2Text models have a very similar amount of parameters (117M).

\section{Model hyper-parameters}
\label{app:hyperparameters}
We keep Logic2Text's hyper-parameters the same as \citet{Chen2020a}. We refer the reader to the paper. Regarding the Table2Logic model in  \tlt, which is based on \citet{Brunner2021}'s Valuenet, we changed the grammar and added additional input data, as well as changing the code accordingly to our use case. We use the same hyper-parameters as stated in the paper, with the exception of the base learning rate, beam size, number epochs, and gradient clipping. This is the list of hyper-parameters used by Table2Logic for the model \tlt:
\vspace{0.4cm}

\resizebox{\textwidth}{!}{
\begin{tabular}{ll}
    \textbf{Random seed}: 90 & \textbf{Attention vector size}: 300 \\
    \textbf{Maximum sequence lengthy}: 512 & \textbf{Grammar type embedding size}: 128 \\
    \textbf{Batch size}: 8 & \textbf{Grammar node embedding size}: 128 \\
    \textbf{Epochs}: 50 & \textbf{Column node embedding size}: 300 \\
    \textbf{Base learning rate}: $5*10^{-5}$ & \textbf{Index node embedding size}: 300 \\
    \textbf{Connection learning rate}: $1*10^{-4}$ & \textbf{Readout}: 'identity' \\
    \textbf{Transformer learning rate}: $2*10^{-5}$ & \textbf{Column attention}: 'affine' \\
    \textbf{Scheduler gamma}: 0.5 & \textbf{Dropout rate}: 0.3 \\
    \textbf{ADAM maximum gradient norm}: 1.0 & \textbf{Largest index for I nodes}: 20 \\
    \textbf{Gradient clipping}: 0.1 & \textbf{Include OOV token}: True \\
    \textbf{Loss epoch threshold}: 50 & \textbf{Beam size}: 2048 \\
    \textbf{Sketch loss weight}: 1.0 & \textbf{Max decoding steps}: 50 \\
    \textbf{Word embedding size}: 300 & \textbf{False Candidate Rejection}: True \\
    \textbf{Size of LSTM hidden states}: 300 & \\
\end{tabular}}
\clearpage
\section{Logical Form grammar}
\label{app:lf-grammar}
\begin{figure}[ht]
\centering
\includegraphics[width=14cm]{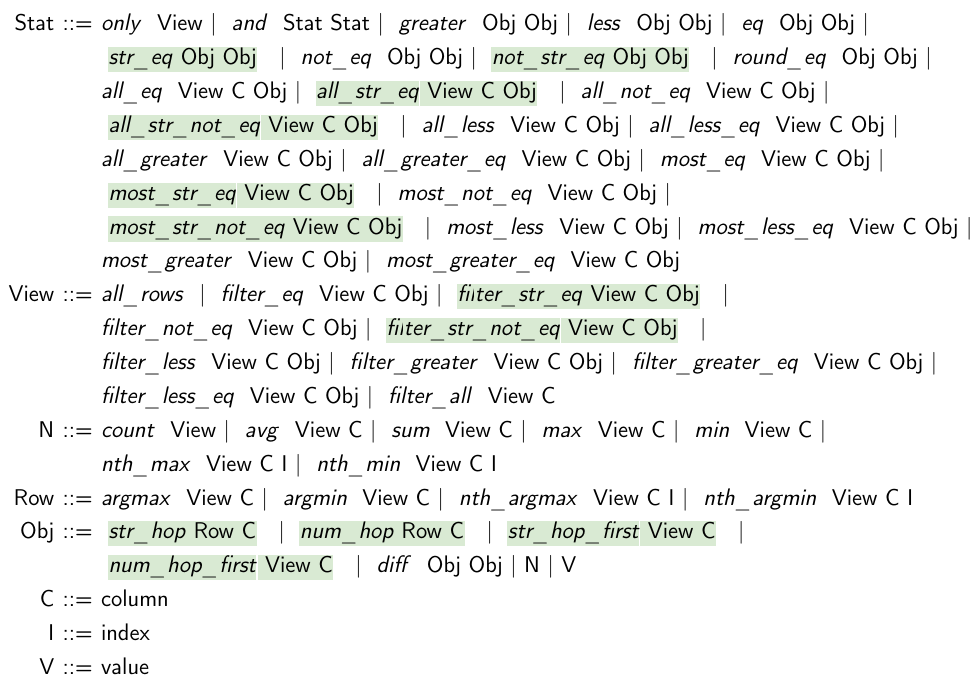}
\caption{The logical form Grammar after fixing the ambiguity issues in the original version \citep{Chen2020a}. We follow the same notation as in IRNet and Valuenet. The tokens to the left of the ::= represent non-terminals (node types in the graph). Tokens in italics represent the possible rules for each node, with pipes ($|$) separating the rules. The rules added to the original grammar in order to fix ambiguity issues are highlighted in green. }
\label{tab:grammar}
\end{figure}

\section{Logic2text errors}
\label{app:qualitative-logic-to-text-examples}
This section shows examples of error cases where the logic-to-text stage of the pipeline failed to produce faithful sentences given a gold LF. We include one example for each error type, including table, caption, gold logical form and generated description. See Section \ref{subsec:qualitative-logic-to-text} for more details.

\begin{figure}[ht]
\small
\flushleft
\subsection{Comparative arithmetic}
\label{subapp:comparative-arithmetic}
\vspace{0.2cm}

\textbf{Caption:} fil world luge championships 1961
\vspace{0.4cm}

\textbf{Table:}
\vspace{0.2cm}

\def\arraystretch{1.2}
\setlength\tabcolsep{2.0pt}
\begin{tabular}{|l|l|l|l|l|l|} 
\hline
\textbf{rank} & \textbf{nation} & \textbf{gold} & \textbf{silver} & \textbf{bronze} & \textbf{total}  \\ 
\hline
1             & austria         & 0             & 0               & 3               & 3               \\
2             & italy           & 1             & 1               & 0               & 2               \\
3             & west germany    & 0             & 2               & 0               & 2               \\
4             & poland          & 1             & 0               & 0               & 1               \\
5             & switzerland     & 1             & 0               & 0               & 1               \\
\hline
\end{tabular}
\vspace{0.4cm}

\textbf{Logical Form:} 
\vspace{0.2cm}

\includegraphics[width=40mm]{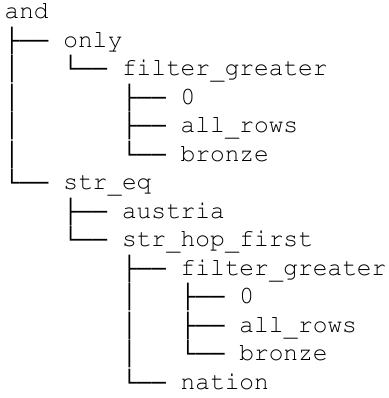}

\vspace{0.4cm}
\textbf{\tlt~ sentence:} austria was the only country to win 0 bronze medals at the fil world luge championships .
\vspace{0.2cm}

\textbf{Gold sentence:} austria was the only country to have bronze medals in the luge championship in 1961 .

\end{figure}

\begin{figure}[ht]
\small
\flushleft
\subsection{LF omission}
\label{subapp:lf-fidelity}
\vspace{0.2cm}

\textbf{Caption:} geography of moldova
\vspace{0.4cm}

\textbf{Table:}
\vspace{0.2cm}

\def\arraystretch{1.2}
\setlength\tabcolsep{2.0pt}

\resizebox{\textwidth}{!}{
\begin{tabular}{|l|l|l|l|l|}
\hline
          \textbf{land formation} & \textbf{area , km square} & \textbf{of which currently forests , km square} & \% \textbf{forests} &  \textbf{habitat type} \\
\hline
northern moldavian hills &             4630 &                                    476 &    10.3 \% & forest steppe \\
   dniester - rāut ridge &             2480 &                                    363 &    14.6 \% & forest steppe \\
      middle prut valley &             2930 &                                    312 &    10.6 \% & forest steppe \\
            bălţi steppe &             1920 &                                     51 &     2.7 \% &        steppe \\
  ciuluc - soloneţ hills &             1690 &                                    169 &    10.0 \% & forest steppe \\
corneşti hills ( codru ) &             4740 &                                   1300 &    27.5 \% &        forest \\
    lower dniester hills &             3040 &                                    371 &    12.2 \% & forest steppe \\
       lower prut valley &             1810 &                                    144 &     8.0 \% & forest steppe \\
           tigheci hills &             3550 &                                    533 &    15.0 \% & forest steppe \\
            bugeac plain &             3210 &                                    195 &     6.1 \% &        steppe \\
part of podolian plateau &             1920 &                                    175 &     9.1 \% & forest steppe \\
 part of eurasian steppe &             1920 &                                    140 &     7.3 \% &        steppe \\
 \hline
\end{tabular}}
\vspace{0.4cm}

\textbf{Logical Form:} 
\vspace{0.2cm}

\includegraphics[width=40mm]{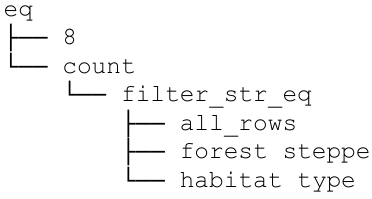}

\vspace{0.4cm}
\textbf{\tlt~ sentence:} there are 8 habitats that can be found in moldova .
\vspace{0.2cm}

\textbf{Gold sentence:} 8 land formations are classified with a habitat type of forest steppe .

\end{figure}

\begin{figure}[ht]
\small
\flushleft
\subsection{Verbalization}
\label{subapp:verbalization}
\vspace{0.2cm}

\textbf{Caption:} seattle supersonics all - time roster
\vspace{0.4cm}

\textbf{Table:}
\vspace{0.2cm}

\def\arraystretch{1.2}
\setlength\tabcolsep{2.0pt}
\resizebox{\textwidth}{!}{
\begin{tabular}{|l|l|l|l|l|l|}
\hline
         \textbf{player} &   \textbf{nationality} & \textbf{jersey number ( s )} & \textbf{position} &                   \textbf{years} &             \textbf{from} \\
\hline
     craig ehlo & united states &                   3 &       sg &             1996 - 1997 & washington state \\
     dale ellis & united states &                   3 &  sg / sf & 1986 - 1991 1997 - 1999 &        tennessee \\
 pervis ellison & united states &                  29 &        c &                    2000 &       louisville \\
francisco elson &   netherlands &                  16 &        c &                    2008 &       california \\
   reggie evans & united states &             34 , 30 &       pf &             2002 - 2006 &             iowa \\
  patrick ewing & united states &                  33 &   center &             2000 - 2001 &       georgetown \\
\hline
\end{tabular}}
\vspace{0.4cm}

\textbf{Logical Form:} 
\vspace{0.2cm}

\includegraphics[width=43mm]{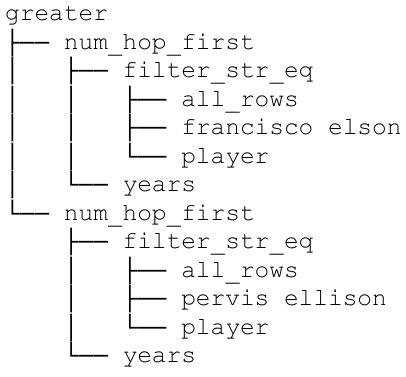}

\vspace{0.4cm}
\textbf{\tlt~ sentence:} foulisco elson played for the supersonics after pervis ellison .
\vspace{0.2cm}

\textbf{Gold sentence:} francisco elson played 8 years later thanpervis ellison .

\end{figure}

\clearpage

\section{Examples of faithful \tlt~ sentences where LF is different to gold}
\label{app:faithful-diff-lf-full-examples}
This section shows examples of automatic LFs from \tlt~ that resulted in faithful sentences in the manual evaluation while being different from their gold LF references. Each example extends the information shown in Table \ref{tab:qualitative_no-gold-faithful}.

\begin{figure}[ht]
\small
\flushleft
\subsection{Similar structure, semantically equivalent}
\label{subapp:similar-equivalent}
\vspace{0.2cm}

\textbf{Caption:} list of appalachian regional commission counties
\vspace{0.4cm}

\textbf{Table:}
\vspace{0.2cm}

\def\arraystretch{1.2}
\setlength\tabcolsep{2.0pt}
\resizebox{\textwidth}{!}{
\begin{tabular}{|l|l|l|l|l|l|}
\hline
     \textbf{county} & \textbf{population} & \textbf{unemployment rate} & \textbf{market income per capita} & \textbf{poverty rate} &       \textbf{status} \\
\hline
   allegany &      49927 &             5.8 \% &                    16850 &       15.5 \% &       - risk \\
     broome &     200536 &             5.0 \% &                    24199 &       12.8 \% & transitional \\
cattaraugus &      83955 &             5.5 \% &                    21285 &       13.7 \% & transitional \\
 chautauqua &     136409 &             4.9 \% &                    19622 &       13.8 \% & transitional \\
    chemung &      91070 &             5.1 \% &                    22513 &       13.0 \% & transitional \\
   chenango &      51401 &             5.5 \% &                    20896 &       14.4 \% & transitional \\
   cortland &      48599 &             5.7 \% &                    21134 &       15.5 \% & transitional \\
   delaware &      48055 &             4.9 \% &                    21160 &       12.9 \% & transitional \\
     otsego &      61676 &             4.9 \% &                    21819 &       14.9 \% & transitional \\
  schoharie &      31582 &             6.0 \% &                    23145 &       11.4 \% & transitional \\
   schuyler &      19224 &             5.4 \% &                    21042 &       11.8 \% & transitional \\
    steuben &      98726 &             5.6 \% &                    28065 &       13.2 \% & transitional \\
      tioga &      51784 &             4.8 \% &                    24885 &        8.4 \% & transitional \\
\hline
\end{tabular}}
\vspace{0.4cm}

\textbf{\tlt~ Logical Form:} 
\vspace{0.2cm}

\includegraphics[width=49mm]{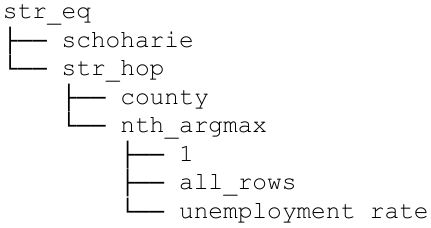}
\vspace{0.4cm}

\textbf{Gold Logical Form:} 
\vspace{0.2cm}

\includegraphics[width=49mm]{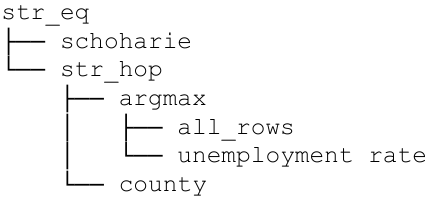}

\vspace{0.4cm}
\textbf{\tlt~ sentence:} in the list of appalachian regional commission counties , schoharie has the highest unemployment rate .
\vspace{0.2cm}

\textbf{Human sentence:} the appalachian county that has the highest unemployment rate is schoharie .

\end{figure}

\begin{figure}[ht]
\small
\flushleft
\subsection{Similar structure, semantically different}
\label{subapp:similar-different}
\vspace{0.2cm}

\textbf{Caption:} dick rathmann
\vspace{0.4cm}

\textbf{Table:}
\vspace{0.2cm}

\def\arraystretch{1.2}
\setlength\tabcolsep{2.0pt}
\begin{tabular}{|l|l|l|l|l|}
\hline
\textbf{year} &    \textbf{qual} &     \textbf{rank} &   \textbf{finish} &     \textbf{laps} \\
\hline
1950 & 130.928 &       17 &       32 &       25 \\
1956 & 144.471 &        6 &        5 &      200 \\
1957 & 140.780 & withdrew & withdrew & withdrew \\
1958 & 145.974 &        1 &       27 &        0 \\
1959 & 144.248 &        5 &       20 &      150 \\
1960 & 145.543 &        6 &       31 &       42 \\
1961 & 146.033 &        8 &       13 &      164 \\
1962 & 147.161 &       13 &       24 &       51 \\
1963 & 149.130 &       14 &       10 &      200 \\
1964 & 151.860 &       17 &        7 &      197 \\
\hline
\end{tabular}
\vspace{0.4cm}

\textbf{\tlt~ Logical Form:} 
\vspace{0.2cm}

\includegraphics[width=38mm]{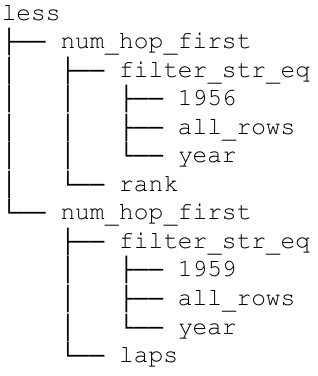}
\vspace{0.4cm}

\textbf{Gold Logical Form:} 
\vspace{0.2cm}

\includegraphics[width=38mm]{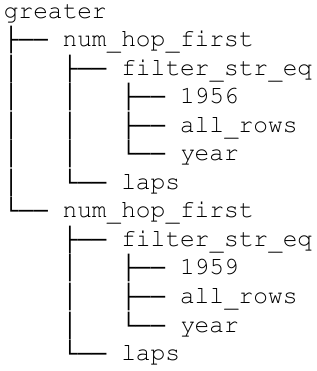}

\vspace{0.4cm}
\textbf{\tlt~ sentence:} dick rathmann had a lower rank in 1956 than he did in 1959 .
\vspace{0.2cm}

\textbf{Human sentence:} dick rathmann completed more laps in the indianapolis 500 in 1956 than in 1959 .

\end{figure}

\begin{figure}[ht]
\small
\flushleft
\subsection{Different structure, semantically different}
\label{subapp:different-different}
\vspace{0.2cm}

\textbf{Caption:} 2005 houston astros season
\vspace{0.4cm}

\textbf{Table:}
\vspace{0.2cm}

\def\arraystretch{1.2}
\setlength\tabcolsep{2.0pt}
\resizebox{\textwidth}{!}{
\begin{tabular}{|l|l|l|l|l|l|l|}
\hline
   \textbf{date} & \textbf{winning team} &  \textbf{score} & \textbf{winning pitcher} &    \textbf{losing pitcher} & \textbf{attendance} &  \textbf{location} \\
\hline
 may 20 &        texas &  7 - 3 &    kenny rogers &     brandon backe &      38109 & arlington \\
 may 21 &        texas & 18 - 3 &     chris young &  ezequiel astacio &      35781 & arlington \\
 may 22 &        texas &  2 - 0 &    chan ho park &        roy oswalt &      40583 & arlington \\
june 24 &      houston &  5 - 2 &      roy oswalt & ricardo rodriguez &      36199 &   houston \\
june 25 &        texas &  6 - 5 &     chris young &     brandon backe &      41868 &   houston \\
\hline
\end{tabular}}
\vspace{0.4cm}

\textbf{\tlt~ Logical Form:} 
\vspace{0.2cm}

\includegraphics[width=24mm]{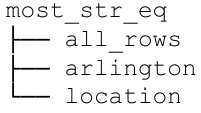}
\vspace{0.4cm}

\textbf{Gold Logical Form:} 
\vspace{0.2cm}

\includegraphics[width=35mm]{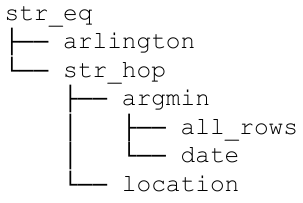}

\vspace{0.4cm}
\textbf{\tlt~ sentence:} most of the games of the 2005 houston astros ' season were played in the location of arlington .
\vspace{0.2cm}

\textbf{Human sentence:} arlington was the first location used in the 2005 houston astros season .

\end{figure}

\begin{figure}[ht]
\small
\flushleft
\subsection{Simpler, more informative semantic}
\label{subapp:simpler}
\vspace{0.2cm}

\textbf{Caption:} 2006 asp world tour
\vspace{0.4cm}

\textbf{Table:}
\vspace{0.2cm}

\def\arraystretch{1.2}
\setlength\tabcolsep{2.0pt}
\resizebox{\textwidth}{!}{
\begin{tabular}{|l|l|l|l|l|}
\hline
             \textbf{location} &          \textbf{country} &                      \textbf{event} &                        \textbf{winner} &                   \textbf{runner - up} \\
\hline
           gold coast &        australia &        roxy pro gold coast & melanie redman - carr ( aus ) &        layne beachley ( aus ) \\
              tavarua &             fiji &              roxy pro fiji & melanie redman - carr ( aus ) &        layne beachley ( aus ) \\
    teahupoo , tahiti & french polynesia & billabong pro tahiti women & melanie redman - carr ( aus ) &     chelsea georgeson ( aus ) \\
              itacarã &           brazil &        billabong girls pro &        layne beachley ( aus ) &    jessi miley - dyer ( aus ) \\
             hossegor &           france &  rip curl pro mademoiselle &     chelsea georgeson ( aus ) & melanie redman - carr ( aus ) \\
          manly beach &        australia & havaianas beachley classic &     stephanie gilmore ( aus ) &        layne beachley ( aus ) \\
sunset beach , hawaii &    united states &                   roxy pro &       melanie bartels ( haw ) &     stephanie gilmore ( aus ) \\
 honolua bay , hawaii &    united states &              billabong pro &    jessi miley - dyer ( aus ) &        keala kennelly ( haw ) \\
\hline
\end{tabular}}
\vspace{0.4cm}

\textbf{\tlt~ Logical Form:} 
\vspace{0.2cm}

\includegraphics[width=37mm]{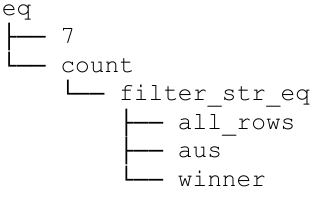}
\vspace{0.4cm}

\textbf{Gold Logical Form:} 
\vspace{0.2cm}

\includegraphics[width=40mm]{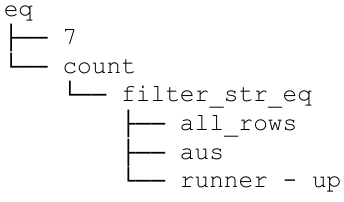}

\vspace{0.4cm}
\textbf{\tlt~ sentence:} aus won 7 events in the 2006 asp world tour .
\vspace{0.2cm}

\textbf{Human sentence:} seven of the individuals that were the runner up were from aus .

\end{figure}

\end{document}